\documentclass{article}
\pdfoutput=1
\usepackage[preprint]{spconf}
\usepackage{amsmath,graphicx}
\usepackage{amsfonts}
\usepackage[ruled]{algorithm2e}
\usepackage{algorithmic,algorithm2e}
\usepackage{booktabs}
\usepackage{multirow}
\usepackage{color}
\usepackage{floatrow}
\usepackage{marvosym}

\copyrightnotice{\begin{minipage}{\textwidth}
  \centering
  \copyright\ IEEE 2021. Personal use of this material is permitted. Permission from IEEE must be obtained for all other uses, in any current or future media, including reprinting/republishing this material for advertising or promotional purposes, creating new collective works, for resale or redistribution to servers or lists, or reuse of any copyrighted component of this work in other works.
  \end{minipage}
  }

\newcommand{\czd}{\color{black}}

\title{Spatio-Temporal Graph Complementary Scattering Networks }
\name{Zida~Cheng, Siheng~Chen\textsuperscript{\Letter}, Ya~Zhang\textsuperscript{\Letter} }

\address{Cooperative Medianet Innovation Center, Shanghai Jiao Tong University, Shanghai, China}
\begin{document}
% \ninept
%
\maketitle
\begin{abstract}
Spatio-temporal graph signal analysis has a significant impact on a wide range of applications, including hand/body pose action recognition. To achieve effective analysis, spatio-temporal graph convolutional networks (ST-GCN) leverage the powerful learning ability to achieve great empirical successes; however, those methods need a huge amount of high-quality training data and lack theoretical interpretation. To address this issue, the spatio-temporal graph scattering transform (ST-GST) was proposed to put forth a theoretically interpretable framework; however, the empirical performance of this approach is constrainted by the fully mathematical design. To benefit from both sides, this work proposes a novel complementary mechanism to organically combine the spatio-temporal graph scattering transform and neural networks, resulting in the proposed~\emph{spatio-temporal graph complementary scattering networks} (ST-GCSN). The essence is to leverage the mathematically designed graph wavelets with pruning techniques to cover major information and use trainable networks to capture complementary information. The empirical experiments on hand pose action recognition show that the proposed ST-GCSN outperforms both ST-GCN and ST-GST.
\end{abstract}
\begin{keywords}
Spatio-temporal graph, graph scattering network, complementary, action recognition
\end{keywords}
\section{Introduction}
\label{sec:intro}

\begin{figure}[t!]
    \centering
    \includegraphics[width=8cm]{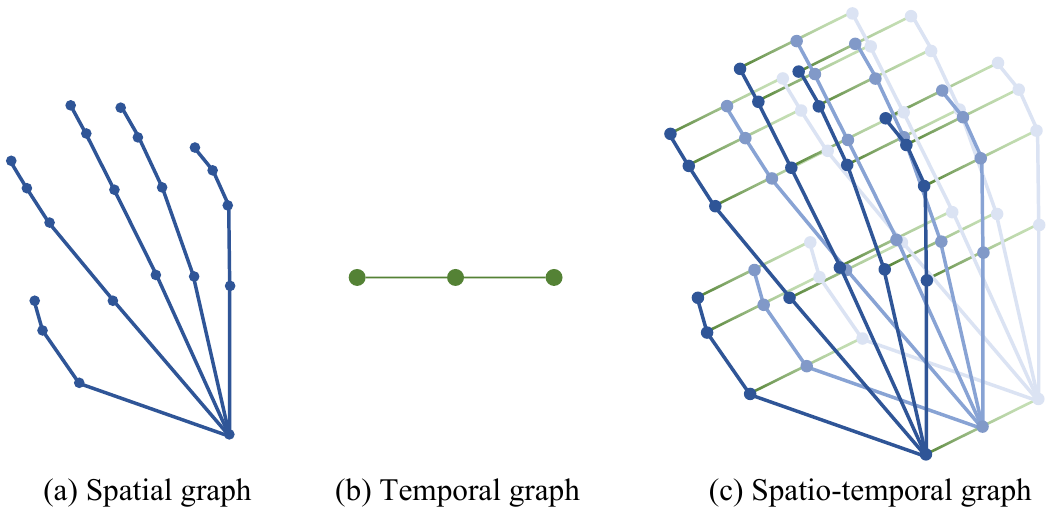}
    \caption{A spatio-temporal graph representing hand pose. The vertices in temporal graph are sequentially connected.}
    \label{fig:handsequence}
\end{figure}

Spatio-temporal graph is an effective tool for modeling dynamic non-Euclidean data. Spatio-temporal graph signal analysis is widely useful in many applications, e.g., hand/body pose action recognition~\cite{stgcn,agcn,LMS_recog_2019,recog_icassp,hand_icassp} and prediction~\cite{LMS_recog_pred_2021,LMS_pred_2021,body_pred_mao}, multi-agent trajectory prediction~\cite{Hu_2020_CVPR,traj_pred_yu} as well as traffic flow forecasting~\cite{traffic_fore_zhang, forcasting_shin}. Figure~{\ref{fig:handsequence}} shows a spatio-temporal graph of the hand pose.

To analyze spatio-temporal graph signals, spatio-temporal graph convolutional networks~(ST-GCN) is emerging as a powerful learning-based method and has achieved great empirical success~\cite{stgcn,agcn}. {\czd However, the design of network architecture lacks theoretical interpretation}. It is thus hard to explain the design rationale and further improve the architectures. Furthermore, ST-GCN also needs a large amount of high-quality labeled data, which might not be available in many practical scenarios.  In contrast to ST-GCN, spatio-temporal graph scattering transform~(ST-GST) was proposed to provide a mathematically interpretable framework~\cite{stgst}. It iteratively applies mathematically designed spatio-temporal graph wavelets and nonlinear activation functions on the graph signal. Since the filter banks do not need any training, ST-GST still works well with limited data. ST-GST also assures some nice theoretical properties, e.g., stability to perturbation of graph signals. However, the empirical performance of  ST-GST is constrained by the nontrainable framework, especially when sufficient data are accessible.

In this work, we propose novel \emph{spatio-temporal graph complementary scattering networks}~(ST-GCSN), which organically combines the main architecture of ST-GST and the learning ability of ST-GCN. Our design rationale is to leverage the mathematically designed graph wavelets with pruning techniques to cover major information and use trainable graph convolutions to capture complementary information. The core of ST-GCSN is new~\emph{graph complementary scattering layer}. In each original graph scattering layer of ST-GST, we use the pruning techniques~\cite{prune} to preserve major branches and prune non-informative ones; we next replace the pruned branches with trainable graph convolutions. Those trainable branches would adaptively capture complementary information missed by the preserved major branches. Therefore, the proposed graph complementary scattering layer has more learning ability than the original graph scattering layer; meanwhile, the proposed graph complementary scattering layer follows mathematically designed graph wavelets~\cite{wavelets,diffusion_st} and has a more clear design rationale than purely trainable graph convolution layer. We conduct experiments for hand pose action recognition task on FPHA~\cite{FPHA} dataset and the results show that our method improves recognition accuracy significantly compared with ST-GCN and ST-GST.

Our main contributions are as follows:

$\bullet$ We propose a novel spatio-temporal graph complementary scattering network. It organically combines mathematically designed ST-GST and trainable ST-GCN, achieving  both theoretical interpretability and learning ability. 

$\bullet$ We propose a novel graph complementary scattering layer~(GCSL) as the basic block, which leverages mathematically designed graph wavelets with pruning techniques to cover major information and uses trainable graph convolutional layer to capture complementary information.

$\bullet$ Extensive experiments are conducted on FPHA dataset for hand pose action recognition task. The proposed method improve classification accuracy by 2.43\% / 1.56\% compared with ST-GCN/ST-GST. Ablation studies show GCSLs function as complementary component as expected.

\section{Preliminary}
\label{sec:method}

\subsection{Spatio-temporal graph scattering transform}
ST-GST~\cite{stgst} iteratively applys spatio-temporal graph wavelets and nonlinear activation, acquiring a tree-like structure. Each node in the tree is a spatio-temporal graph signal.

\begin{figure*}[t!]
    \centering
    \includegraphics[width=17.5cm]{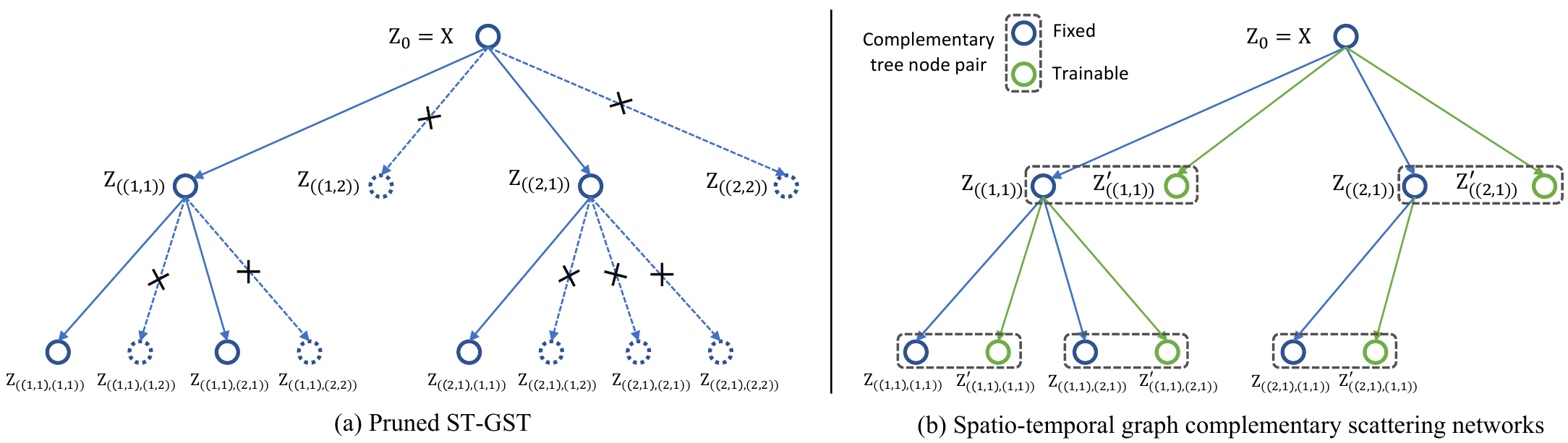}
    \caption{(a) Pruned scattering tree of ST-GST with 2 layer, $J_s = J_t = 2$. Dotted tree nodes are pruned and the children tree nodes of the pruned ones are omitted. (b) Architecture of the proposed spatio-temporal graph complementary scattering network. Each fixed tree node (in blue) in the pruned scattering tree is complemented by a trainable tree node (in green). }
    \label{fig:main img}
\end{figure*}

With the hand pose in Figure~\ref{fig:handsequence} as an example, the hand has $ N $ joints and the action sequence has $ T $ time steps. We see it as a spatio-temporal graph with $ N $ spatial vertices and $ T $ temporal vertices. The input graph signal is $ \mathbf{X} \in \mathbb{R}^{N \times T} $, a general spatio-temporal graph filter is be defined as
\begin{equation*}
\mathbf{H}\left(\mathbf{S}_{s}\right) \mathbf{X} \mathbf{G}^{\top}\left(\mathbf{S}_{t}\right) = \left(\sum_{p=0}^{P-1} h_{p} \mathbf{S}_{s}^{p}\right) \mathbf{X}\left(\sum_{q=0}^{Q-1} g_{q} \mathbf{S}_{t}^{q}\right)^{\top},
\label{eq:graph filter}
\end{equation*}
where $\mathbf{S}_{s}$ and $ \mathbf{S}_{t} $ are the spatial and temporal graph shift matrices, respectively. The spatial and temporal graph filters, $\mathbf{H}\left(\mathbf{S}_{s}\right) $ and  $ \mathbf{G}^{\top}\left(\mathbf{S}_{t}\right)$ are the polynomial function of graph shift matrices. Then we can design spatial and temporal graph wavelets. Here we use the lazy random walk matrix as the graph shift:  $\mathbf{S}_{s}$ = $\mathbf{P}_{s}$ = $\frac{1}{2} ( \mathbf{I} + \mathbf{D}_s^{-1} \mathbf{A}_s  )$, where $ \mathbf{A}_s $ is the adjacent matrix in spatial domain and $\mathbf{D}_s$ is the diagonal degree matrix; similarly, we use $\mathbf{P}_{t}$ as the time-domain shift. The $\mathbf{P}$ matrices are Markov matrices and row sums are all one, which can represent a signal diffusion process over a graph. The commonly used graph wavelets in ST-GST are defined as

\begin{equation*}
\begin{aligned}
\{\mathbf{H}_{j_{1}}\left(\mathbf{P}_{s}\right) &= \mathbf{P}_{s}^{2^{j_1 - 1}} -  \mathbf{P}_{s}^{2^{j_1}} \}_{j_{1}=1}^{J_{s}}, \\ \{\mathbf{G}_{j_{2}}\left(\mathbf{P}_{t}\right) &= \mathbf{P}_{t}^{2^{j_2 - 1}} -  \mathbf{P}_{t}^{2^{j_2}} \}_{j_{2}=1}^{J_{t}},
\end{aligned}
\label{eq:wavelets}
\end{equation*}
where $j_1$ / $j_2$ are the spatial/temporal scale. $J_s$ and $ J_t $ are space/time scale numbers.

Let $\mathbf{Z} \in \mathbb{R}^{N \times T} $ be a spatio-temporal graph signal input. Each spatio-temporal filter is composed of a spatial filter chosen from $ \left\{\mathbf{H}_{j_{1}} \right\}_{j_{1}=1}^{J_{s}} $ and a temporal filter from $ \left\{\mathbf{G}_{j_{2}} \right\}_{j_{2}=1}^{J_{t}} $. Thus we can define $ J = J_s \times J_t $ filters and get $J$ output signals by convolving them with $\mathbf{Z}$. 
% Then the nonlinear activation $\sigma$ is applied. Usually the absolute value function is used which has the property of energy-preserving. 
Finally, each output signal is $\mathbf{Z}_{\left(j_{1}, j_{2}\right)}=\sigma\left(\mathbf{H}_{j_{1}}\left(\mathbf{P}_{s}\right) \mathbf{Z} \mathbf{G}_{j_{2}}^{\top}\left(\mathbf{P}_{t}\right)\right)$. $\sigma$ is the absolution value function for nonlinear activation. With spatio-temporal graph signals seen as tree nodes, a parent tree node $\mathbf{Z}$ generates $J$ children tree nodes in one scattering layer.

The full ST-GST framework is shown in Figure~\ref{fig:main img}(a). $\mathbf{Z}_0 = \mathbf{X}$ is the root node. We iteratively apply scattering layers and the tree grow exponentially with base of $J$. Each tree node can be indexed by the path from root to it. For example, $p^{(\ell)}=\left(j_{1}^{(1)}, j_{2}^{(1)}\right), \ldots,\left(j_{1}^{(\ell)}, j_{2}^{(\ell)}\right)$ is the path from root to a node in the $\ell$-th layer, denoted as $\mathbf{Z}_{(p^{(\ell)})}$. Finally, the graph signals across all the tree nodes are concatenated to form the final scattering feature, which is used for downstream tasks. ST-GST has nice theoretical properties, such as the stability to the graph signal and structural perturbations.

\subsection{Pruned graph scattering transform}
\label{sec:pruning}
The complexity of ST-GST grows exponentially with layer number. To reduce complexity, \cite{prune} prunes the scattering tree according to the ratio of a child node energy to the parent node energy. We can easily extend it to the spatio-temporal graph domain. For a parent node $\mathbf{Z}_{p} $ and a child node $\mathbf{Z}_{c}$, $\mathbf{Z}_{c}$~(and its children) are pruned if $ \frac{|| \mathbf{Z}_{c} || }{||\mathbf{Z}_{p}||} < \tau $, where $||\cdot||$ means Frobenius norm and $\tau$ is a user-specific threshold. Figure~\ref{fig:main img}(a) shows an example of pruned scattering tree.~\cite{prune} shows that this pruning design can still preserve the scattering's stability property.

\section{Spatio-temporal graph complementary scattering network}
We first introduce a trainable graph complementary scattering layer as the basic block of ST-GCSN.  We then present agent-parameter training, which enables the training of ST-GCSN.

\subsection{Graph complementary scattering layer}

\textbf{Scattering layer with pruning.}
The idea is to use a pruned spatio-temporal graph scattering layer to be the initial structure of a graph complementary scattering layer. For an input spatio-temporal graph signal $\mathbf{Z}_{(p^{(\ell)})}$, an ordinary spatio-temporal graph scattering layer outputs:
\begin{equation*}
\begin{aligned}
\mathbf{Z}_{ \left( p^{(\ell)}, (j_{1},j_{2}) \right) }&=\sigma\left(\mathbf{H}_{j_{1}}\left(\mathbf{P}_{s}\right) \mathbf{Z}_{(p^{(\ell)})} \mathbf{G}_{j_{2}}^{\top}\left(\mathbf{P}_{t}\right)\right),\\ j_1 = &1,...,J_s \quad j_2=1,...J_t,
\end{aligned}
\label{eq:ori filter}
\end{equation*}
That means each input parent tree node generates $J=J_s \times J_t$ children tree nodes. We then perform the pruning technique in Section~\ref{sec:pruning} with threshold $\tau$, the preserved tree node set is:
\begin{equation*}
\begin{aligned}
\mathcal{Z}_f = \{ \mathbf{Z}_{ \left( p^{(\ell)}, (j_{1},j_{2}) \right)} \ &| \  \frac{|| \mathbf{Z}_{ \left( p^{(\ell)}, (j_{1},j_{2}) \right)} || }{||\mathbf{Z}_{(p^{(\ell)})}||} > \tau,\\ j_1 = 1,...,J_s,&\  j_2=1,...J_t \}.
\end{aligned}
\label{eq:preserved node set}
\end{equation*}
This forms the fixed component in a graph complementary scattering layer. The process is shown in Figure~\ref{fig:main img}(a).

\textbf{Complementary tree nodes.} The idea is to construct trainable tree nodes to replace the pruned tree nodes, adaptively complementing to the preserved tree node set in each graph complementary scattering layer; see Figure~\ref{fig:main img}(b). For each preserved tree node $\mathbf{Z}_{ \left( p^{(\ell)}, (j_{1},j_{2}) \right)} \in \mathcal{Z}_f$, we construct its complementary tree node $\mathbf{Z}^{\prime}_{ \left( p^{(\ell)}, (j_{1},j_{2}) \right)}$ . These two tree nodes form a complementary pair:
\begin{equation}
\begin{aligned}
\mathbf{Z}_{ \left( p^{(\ell)}, (j_{1},j_{2}) \right) }&=\sigma\left(\mathbf{H}_{j_{1}}\left(\mathbf{P}_{s}\right) \mathbf{Z}_{(p^{(\ell)})} \mathbf{G}_{j_{2}}^{\top}\left(\mathbf{P}_{t}\right)\right),\\ \mathbf{Z}^{\prime}_{ \left( p^{(\ell)}, (j_{1},j_{2}) \right) }&=\sigma\left( \left(\mathbf{I} - \mathbf{H}_{j_{1}}\left(\mathbf{P}^{\prime}_{s}\right)\right) \mathbf{Z}_{(p^{(\ell)})} \left( \mathbf{I} -  \mathbf{G}_{j_{2}}^{\top}\left(\mathbf{P}^{\prime}_{t}\right)\right)\right),
\end{aligned}
\label{eq:node pair def}
\end{equation}
where $\mathbf{P}^{\prime}_{s}$ and $\mathbf{P}^{\prime}_{t}$ are trainable through backpropagation and initialized as $\mathbf{P}_{s}$ and $\mathbf{P}_{t}$, respectively. We thus obtain 
the trainable tree node set, whose cardinality is the same with the fixed tree node set. This design involves two key-points before training:  (1) the filters of fixed and trainable tree nodes are explicitly complementary. The filter $\mathbf{H}$ in a fixed tree node corresponds to the complementary filter $\mathbf{I}-\mathbf{H}$ in the associated trainable tree node; and (2) all the trainable tree nodes with the same parent share the same $\mathbf{P}^{\prime}_{s}$ and $\mathbf{P}^{\prime}_{t}$, following designing rule of graph wavelets. That gives a more clear design rationale than purely trainable graph convolutional layer.

\textbf{Overall layer.} Finally, a graph complementary scattering layer is formed by both fixed tree nodes and trainable tree nodes. By iteratively apply  graph complementary scattering, we obtain a spatio-temporal graph complementary scattering network, which organically combines the advantages of both ST-GST and ST-GCN; see Figure~\ref{fig:main img}(b). For downstream tasks, all tree nodes are concatenated as the final feature.

\subsection{Agent-parameter training}
To ensure that
the trainable tree nodes in graph complementary scattering layers still retain nice theoretical properties as the fixed nodes, we propose the agent-parameter training mechanism. Recall that we make $\mathbf{P}^{\prime}_{s}$ and $\mathbf{P}^{\prime}_{t}$ trainable, which are Markov matrices with rows normalized. To preserve the row-normalize property, for a Markov matrix $\mathbf{P}^{\prime}$, we do not train it directly but use an agent parameter matrix $\mathbf{M}$ and let: $\mathbf{P}^{\prime} = \mathcal{E}(\mathbf{M})$, where $\mathcal{E}(\cdot)$ means row-wise softmax normalization. We then train $\mathbf{M}$ as the network parameter. 
At the start of training, we initialize $\mathbf{M}$ so that the resultant random walk matrix is equal to the corresponding matrix in the fixed tree nodes: $\mathbf{P}^{\prime}_{init}=\mathcal{E}(\mathbf{M}_{init}) = \mathbf{P}.$

% We can show that when the training process does not change the random walk $\mathbf{P}$ too much,  the proposed spatio-temporal graph complementary scattering network is still stable with respect to graph signal perturbations:

% \textbf{Theorem 1.} A spatio-temporal graph complementary scattering network denoted as $\boldsymbol{\Psi}(\cdot)$ have $L$ layers and $J$ filters each layer. With an input spatio-temporal graph signal $\mathbf{X}$, the network is stable to a bounded perturbation $\mathbf{\Delta}$, in the sense that
% \begin{equation}
% \begin{aligned}
% \frac{\|\boldsymbol{\Psi}(\mathbf{X})-\mathbf{\Psi}(\tilde{\mathbf{X}})\|_{2}}{\sqrt{|\boldsymbol{\Psi}(\mathbf{X})|}} \leq \sqrt{\frac{\sum_{\ell=0}^{L} F_{\ell} B^{2 \ell}}{\sum_{\ell=0}^{L} F_{\ell}}}\|\boldsymbol{\delta}\|_{2},
% \end{aligned}
% \label{eq:theorem}
% \end{equation}
% where $\tilde{\mathbf{X}} = \mathbf{X} + \mathbf{\Delta}$, $F_{\ell}$ is the number of nodes at $\ell$-th layer.

To summarize, we build ST-GCSN by graph complementary scattering layers, which are composed of fixed tree nodes and trainable tree nodes. We use spatio-temporal graph scattering layer and pruning technique to get fixed nodes. Then we construct trainable tree nodes to capture complementary information that fixed tree nodes missed. The adaptive nodes are designed in an explicitly complementary manner and also follow the graph wavelets design rule.

\section{Experimental results}
\label{sec:experiment}
\subsection{Experimental setup}
\textbf{Dataset.} First-Person Hand Action Recognition~(FPHA)~\cite{FPHA} contains 1175 action videos from 45 different action categories. Each video contains a right hand manipulating an object.
The dataset provide 3D coordinate annotation of 21 hand joints for every frame. We clip and pad the skeleton coordinate sequences so that each sequence contains 200 frames. We then uniformly sample 67 frames from each sequence. We use 600 sequences for training and 575 for testing, following the 1:1 setting in \cite{FPHA} . The original data consist of 3 coordinate dimensions $(x,y,z)$, we use them as three independent channels in the root graph signal node.

\textbf{Implementation details.} We use 2 scattering layers with $J_s=20, J_t=5$, resulting in 10101 nodes before pruning. We set pruning threshold $\tau=0.002$, preserving 2693 fixed tree nodes after pruning. We pool the final feature by temporal average to reduce cardinality. The classifier is implemented by a multi-layer perceptron with one hidden layer.

\begin{table}[t!]
\centering
\caption{Hand action recognition accuracy on FPHA. Our method outperms both ST-GST and ST-GCN.}
\label{tab:main}
    \begin{tabular}{cc}
    \toprule
    Method            & Accuracy (\%) \\ \hline
    Transition Forest~\cite{tf} & 80.69         \\
    Huang et al.~\cite{huang}      & 84.35         \\
    ST-GCN~\cite{agcn}            & 86.32         \\
    ST-GST~\cite{stgst}              & 87.19         \\ 
    \textbf{ST-GCSN}    & \textbf{88.75}        \\ 
    \bottomrule
    \end{tabular}
\end{table}

\subsection{Results}

\textbf{Primary results} \  Table \ref{tab:main} reports the action recognition accuracy on FPHA. The proposed ST-GCSN outperforms ST-GST, ST-GCN and other strong baselines. Our method shows superiority over purely trained graph convolutions and fully designed methods.

\textbf{Effect of complementary mechanism} \  The complementary mechanism is the core of ST-GCSN. It involves two aspects: (1) the fixed tree nodes and trainable tree nodes capture complementary information; (2) the filters in fixed tree nodes and trainable tree nodes are in the complementary form $(\mathbf{H}, \mathbf{I-H})$, see Eq.\ref{eq:node pair def}. In Table~\ref{tab:complement}, We conduct ablation experiments for both aspects: (1) We construct two networks using only fixed/trainable tree nodes. Results show that best accuracy is achieved only when fixed and trainable nodes are both utilized. That validates that the fixed and trainable component of ST-GCSN work in a complementary manner. (2) We replace the $(\mathbf{H},\mathbf{I} - \mathbf{H})$ filter design with $(\mathbf{H}, \mathbf{H})$, that is, 
$\mathbf{Z}^{\prime}_{  p^{(\ell)}, (j_{1},j_{2})  }=\sigma\left( \left( \mathbf{H}_{j_{1}}\left(\mathbf{P}^{\prime}_{s}\right)\right) \mathbf{Z}_{(p^{(\ell)})}  \mathbf{G}_{j_{2}}^{\top}\left(\mathbf{P}^{\prime}_{t}\right)\right)$; called~\emph{w/o complementary mechanism}. Accuracy drops significantly without the explicit complementary filters, validating the necessity of design in Eq.~\ref{eq:node pair def}.

\begin{table}[t]
\centering
\caption{Effect of complementary mechanism. Both fixed and trainable tree nodes with the complementary design significantly improves the performance.}
\label{tab:complement}
\begin{tabular}{cc}
\toprule
Method         & Acc. (\%)   \\ \hline
Fixed tree nodes only & 87.19     \\
Trainable tree nodes only & 87.82     \\
w/o complementary mechanism & 86.26        \\
\textbf{complementary mechanism} & \textbf{88.75}        \\
\bottomrule
\end{tabular}
\end{table}

\begin{figure}[t!]
    \centering
    \includegraphics[width=6.5cm]{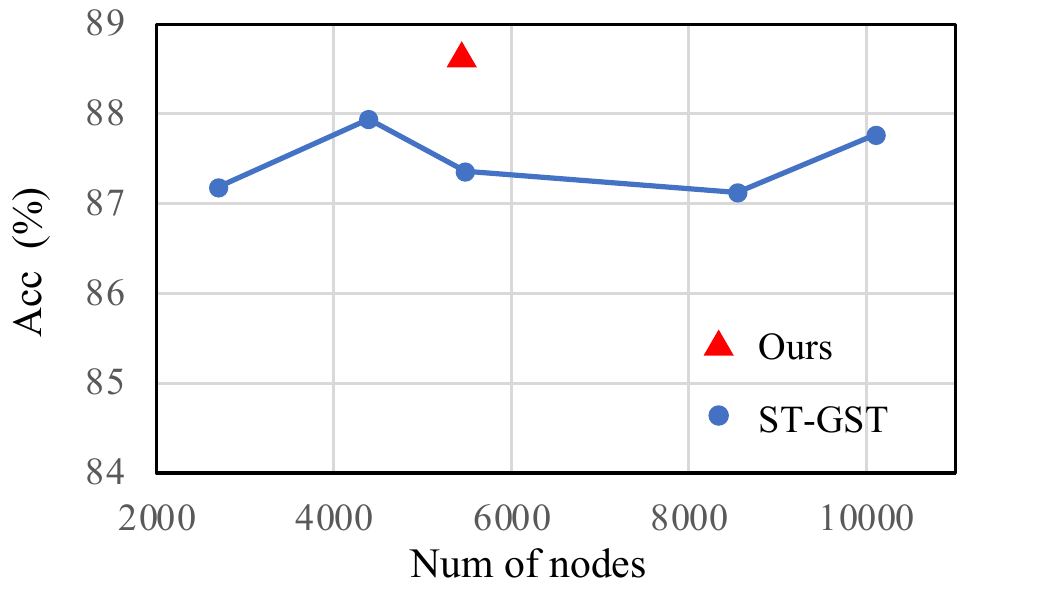}
    \caption{Effect of tree node number. The proposed ST-GCSN outperforms ST-GST with varying number of tree nodes.}
    \label{fig:node num}
\end{figure}

\textbf{Effect of number of tree nodes.}  Compared with ST-GST, the proposed ST-GCSN have both fixed and trainable tree nodes, leading to twice of the tree node cardinality. To analyze the effect of the tree node number, we vary it by applying different pruning thresholds on ST-GST.  Figure~\ref{fig:node num} shows that simply increasing the node number brings little gain. That indicates the improvement of our method is mainly credited to the proposed complementary framework, instead of preserving more tree nodes in ST-GST.

\textbf{Effect of trainability} The proposed graph complementary scattering layer follows the design of graph wavelets, thus acquire more clear design rationale than purely trainable graph convolution. To analyze the effect of trainability, we replace the trainable tree nodes with purely trainable graph convolutional filters.  Table~\ref{tab:trainability} shows that the purely trainable network reaches negligible gain by sacrificing 50 times more trainable parameters.

\begin{table}[t!]
\centering
\caption{Effect of trainability. Proposed ST-GCSN achieves similar performance with 50 times less trainable parameters.}
\label{tab:trainability}
\begin{tabular}{ccc}
\toprule
Method         & Acc. (\%) & Params.   \\ \hline
Purely trainable & 88.99 & 13.27M    \\
\textbf{ST-GCSN} & \textbf{88.75} & 0.23M        \\
\bottomrule
\end{tabular}
\end{table}

\section{Conclusion}
We propose the spatio-temporal graph complementary scattering networks (ST-GCSN) for spatio-temporal graph data analysis. As the basic block of ST-GCSN, each graph complementary scattering layer consists of both mathematically designed tree nodes and trainable tree nodes. With the proposed complementary mechanism, graph complementary scattering layers also have both theoretical interpretability and flexible learning ability. Experimental results validate that the proposed method outperforms both ST-GCN and ST-GST on the task of hand pose action recognition.

\vfill\pagebreak

\bibliographystyle{IEEEbib}
\bibliography{strings,refs}

\end{document}